\documentclass[letterpaper]{article} % DO NOT CHANGE THIS
\usepackage{aaai24}  % DO NOT CHANGE THIS
\usepackage{times}  % DO NOT CHANGE THIS
\usepackage{helvet}  % DO NOT CHANGE THIS
\usepackage{courier}  % DO NOT CHANGE THIS
\usepackage[hyphens]{url}  % DO NOT CHANGE THIS
\usepackage{graphicx} % DO NOT CHANGE THIS
\usepackage{natbib}  % DO NOT CHANGE THIS AND DO NOT ADD ANY OPTIONS TO IT
\usepackage{caption} % DO NOT CHANGE THIS AND DO NOT ADD ANY OPTIONS TO IT
\usepackage{algorithm}
\usepackage{algorithmic}

\usepackage{newfloat}
\usepackage{listings}
\DeclareCaptionStyle{ruled}{labelfont=normalfont,labelsep=colon,strut=off} % DO NOT CHANGE THIS
\floatstyle{ruled}
\newfloat{listing}{tb}{lst}{}
\floatname{listing}{Listing}
\title{\M: Activation-Guided Quantization \\ for Faster Inference of LLMs on the Edge}
\author {
    Xuan Shen\thanks{These authors contributed equally.}\textsuperscript{\rm 1},
    Peiyan Dong\footnotemark[1]\textsuperscript{\rm 1}, 
    Lei Lu\textsuperscript{\rm 1},
    Zhenglun Kong\textsuperscript{\rm 1}, \\
    Zhengang Li\textsuperscript{\rm 1},
    Ming Lin\thanks{Work done before joining Oracle.}\textsuperscript{\rm 2},
    Chao Wu\textsuperscript{\rm 1},
    Yanzhi Wang\textsuperscript{\rm 1} 
}
\affiliations {
    \textsuperscript{\rm 1}Northeastern University \\ %\qquad
    \textsuperscript{\rm 2}Oracle
    
    \{shen.xu, dong.pe, lu.lei1, kong.zhe, li.zhen, cha.wu, yanz.wang\}@northeastern.edu \\
    linming04@gmail \\
}

\usepackage{paralist}
\usepackage[usestackEOL]{stackengine}
\usepackage{bm}
\usepackage{makecell}
\usepackage{amsfonts} 
\usepackage{tabularx}
\usepackage{multirow}
\usepackage{graphicx}
\usepackage{booktabs}
\usepackage{amsfonts,amsmath} 
\usepackage[sort,nocompress]{cite}
\usepackage{tabularray}
\usepackage{url}
\usepackage{pifont}

\newcommand{\M}{Agile-Quant}

\usepackage[dvipsnames]{xcolor}
\newif\ifmodify 
\ifmodify

\newcommand{\xuan}[1]{\textcolor{blue}{#1}}
\newcommand{\py}[1]{\textcolor{Emerald}{#1}}

\else

\newcommand{\xuan}[1]{#1}
\newcommand{\py}[1]{#1}

\fi

\usepackage{bibentry}
\begin{document}

\maketitle

%%%%%%%%% ABSTRACT
\begin{abstract}

Large Language Models (LLMs) stand out for their impressive performance in intricate language modeling tasks.
However, their demanding computational and memory needs pose obstacles for broad use on edge devices.
Quantization is then introduced to boost LLMs' on-device efficiency.
% To fully capitalize on the hardware benefits of model quantization for LLMs, addressing two crucial issues is essential.
% {\large\ding{182}}
Recent works show that 8-bit or lower weight quantization is feasible with minimal impact on end-to-end task performance,
while the activation is still not quantized.
% However, comprehensive literature on the quantization of both weight and activation paradigms for LLMs is currently limited.
% {\large\ding{183}}
On the other hand,
mainstream commodity edge devices still struggle to execute these sub-8-bit quantized networks effectively.
In this paper, we propose \M, an \underline{A}ctivation-\underline{G}uided quantization framework for faster \underline{I}nference of popular \underline{L}arge Language Models (LLMs) on the \underline{E}dge.
% and implement an end-to-end accelerator for multiple edge devices to achieve faster inference.
Considering the hardware profiling and activation analysis, we first introduce a basic activation quantization strategy to balance the trade-off of task performance and real inference speed.
Then we leverage the activation-aware token pruning technique to reduce the outliers and the adverse impact on attentivity.
Ultimately, we utilize the SIMD-based 4-bit multiplier and our efficient TRIP matrix multiplication to implement the end-to-end accelerator for LLMs on multiple edge devices.
We apply our framework on different scales of LLMs including LLaMA, OPT, and BLOOM with 4-bit or 8-bit for the activation and 4-bit for the weight quantization.
% Extensive experiments demonstrate 
Experiments show that \M~achieves simultaneous quantization of model weights and activations while maintaining task performance comparable to existing weight-only quantization methods.
Moreover, in the 8- and 4-bit scenario, \M~achieves an on-device speedup of up to 2.55x compared to its FP16 counterparts across multiple edge devices, marking a pioneering advancement in this domain.
Code: \textcolor{blue}{\url{https://github.com/shawnricecake/agile-quant}}

\end{abstract}

%%%%%%%%% BODY TEXT

\section{Introduction}
% xuan: @peiyan, talk about the applications, say the mix-precision can not be accelerated on mobile devices.

% xuan: revise from introduction, do not talk about outliers, we just claim 2^ with layer-wise (and log2 for the output of softmax) in quantization setting and cite some works

% xuan: pay more attention on token pruning, token pruning can be useful when the input is large, especially for small model which is not strong enough to deal with the large input, focus on small models because we focus on edge devices. The token pruning can be used in the finetuning in future work

% \py{Pre-trained generative models from the Transformer~\cite{vaswani2017attention} family, commonly known as GPT or OPT~\cite{radford2019language,brown2020language,zhang2022opt}, have shown breakthrough performance for complex language modeling tasks, leading to massive academic and practical interest.}
\xuan{
Large Language Models (LLMs)~\cite{touvron2023llama, zhang2022opt, brown2020language, radford2019language, gpt3} based on the Transformer~\cite{vaswani2017attention} family have breakthrough performance in Natural Language Processing (NLP) research area.
}

\textbf{Application Scenarios}.
\py{In real-world decision scenarios, incorporating LLMs inference as a crucial element often necessitates stringent latency requirements. However, one drawback of LLMs is their computational and storage cost, which ranks among the highest for known models. 
Consider GPT3-175B as an example. When stored in a compact float16 format, its parameters require 326GB (in multiples of 1024) of memory. This surpasses the capacity of even the most powerful individual GPUs, not to mention the challenges of running it on hardware-limited edge devices with acceptable latency.
% Two common methods for reducing these overheads are model pruning and quantization, both of which have been extensively studied for deep neural networks (DNNs) in the field of computer vision.
Quantization, in particular, offers a promising approach to substantially improve the inference throughput and energy efficiency of LLMs on edge devices.
This improvement is achieved by harnessing the highly effective 8-bit fixed-point (INT8) operations supported by the SIMD units that are commonly found in edge platforms, such as CPUs and Raspberry Pis.}

% now have many weight-only works
\textbf{Current Limitations}. 
\py{Before fully realizing the on-device benefits of model quantization on LLMs, it's crucial to address two pressing issues that demand careful attention.}
{\large\ding{182}}
\py{Existing works~\cite{frantar-gptq, lin2023awq, xiao2022smoothquant} primarily concentrate on weight-only (4-bit) quantization while leaving activations in the floating-point (FP16) domain.
This approach limits the efficient speed-up of model inference on common edge devices, which typically only support 16x16 and 8x8 integer multipliers. Specifically, activation quantization often has a detrimental effect on task performance, especially when the model size becomes large, due to the emergence of pronounced outliers in activations.
Experiments done by work~\cite{dettmers2022llm} indicate that directly setting these outliers to zero can result in a substantial 45\% degradation in task performance.
Additionally, given the large model size of LLMs, limited academic computing power makes it challenging to afford the associated training costs.
Consequently, Post-Training Quantization (PTQ) has become a prevalent approach, but it falls short of minimizing the quantization error caused by these outliers.
In summary, quantizing the activations of LLMs while handling outliers inside activations is a crucial yet challenging issue.}
{\large\ding{183}}
\py{Mainstream edge processors, such as CPUs and Raspberry Pis, leverage SIMD units to execute multiple operations in parallel efficiently.
SIMD instructions are adept at exploiting byte-level data (8-bit integers) parallelism and are well-supported in common ISAs (Instruction Set Architectures) and DNN processing frameworks.
Examples include GEMMLOWP~\cite{jacob2017gemmlowp} in TensorFlow Lite and QNNPACK~\cite{dukhan2018qnnpack} in PyTorch.
% However, these low-precision libraries prove ineffective for running networks quantized with only 4-bit weights, due to their sole support for byte-level or wider data-parallel execution.
Their low-precision kernels merely zero-extend the sub-byte operands to align them with byte boundaries, treating them as 8-bit or 16-bit operands.} 

% \py{In this paper, we address the above on-device quantization issues while enjoying the powerful performance provided by LLMs.
% Please note that model pruning is also one of the mainstream technologies to accelerate edge computing for neural networks, which is orthogonal to quantization. More hardware benefits can be achieved by combining model pruning and quantization. \todo{why say this? we can wait until the reviewers ask}}

\py{In this paper, we address the above on-device quantization issues while enjoying the powerful performance provided by LLMs.
We propose \M, an activation-guided quantization framework for faster inference of LLMs on the edge.
Specifically, we begin with a fundamental activation quantization strategy based on hardware latency profiling and activation analysis of LLMs, aiming to strike a balance between task performance and on-device inference speed.
We subsequently utilize the activation-aware pruning method to optimize quantization.
This is crucial because quantized tokens often exhibit numerous outliers, causing their attention to shift from the first position to nearby local positions.
By pruning tokens, we effectively eliminate some outliers, as they typically concentrate within the same or adjacent channels of different tokens.
Also, the removal of inattentive tokens can reduce the interaction distance between important tokens.
Finally, we design the edge-oriented optimization for the hardware implementation of \M.
It consists primarily of two components: a SIMD-based 4-bit multiplier to facilitate efficient 4x4 INT4 multiplication, and our efficient Two-Refine Improved by Pruning (TRIP) matrix multiplication designed to mitigate the adverse impact of outliers.}

% summary
% The recent popular LLMs models such as LLaMA~\cite{touvron2023llama}, OPT~\cite{zhang2022opt}, and BLOOM~\cite{scao2022bloom} are adopted to verify the effectiveness of our framework and the efficiency of our method on multiple edge devices.
% We achieve \todo{state-of-the-art} task performance with weight-only works by our proposed framework on those three LLMs.
% % We focus our activation quantization on the matrix multiplications rather than the nonlinear operators such as Softmax, LayerNorm, and ReLU (or SwiGLU in LLaMA), as the matrix multiplication takes the most proportion of inference latency.
% \todo{We also implement several models with different scales on the edge to achieve practical acceleration.
The popular LLMs models such as LLaMA~\cite{touvron2023llama}, OPT~\cite{zhang2022opt}, and BLOOM~\cite{scao2022bloom} are adopted to verify the effectiveness of our framework and the efficiency of our method on multiple edge devices.
\M~can maintain state-of-the-art task performance comparable with weight-only works while achieving practical on-device speedup up to 2.55x.

The contributions of this work are summarized as follows:
\begin{compactitem}
    
    \item We design the activation-guided and edge-oriented quantization strategy for the balance of latency decreasing and task performance.

    \item We design an activation-aware token pruning method to minimize the negative impact on task performance caused by the outliers and the local attentivity.

    \item We propose the SIMD-based 4-bit multiplier and an efficient TRIP matrix multiplication for effective hardware implementation.

    \item We achieve state-of-the-art task performance on several popular datasets with practical on-device speedup.
    
\end{compactitem}

\section{Background and Related Works}
In this section, we first focus on the backgound of post-training quantization for LLMs. 
Then we discuss the low-bit computation on general edge devices.

\subsection{Post-Training Quantization for LLMs}
\py{Post-Training Quantization (PTQ) techniques are widely used for one-shot compressing models, particularly for Large Language Models (LLMs), given the high cost of retraining. These PTQ methods employ accurate solvers to address compression challenges on a per-layer or per-group basis, relying on a limited set of calibration data.
Notably, recent advances in PTQ, like GPTQ~\cite{frantar-gptq}, AWQ~\cite{lin2023awq}, and SpQR~\cite{dettmers2023spqr}, have introduced well-crafted approaches capable of preserving LLM performance effectively. GPTQ leverages second-order information to correct errors, achieving commendable accuracy within a 3-4 bit range. 
%To elaborate, GPTQ quantizes weight matrices column-wise and balances unquantized portions against quantized segments, thereby upholding model performance.
AWQ proposes safeguarding only 1\% of crucial weights to substantially diminish quantization errors.
%They analyze activations instead of weights to identify significant weights for protection, unveiling an inherent link between weights and activations.
SpQR's focus is on reducing quantization to 3-4 bits per parameter for smaller models.
Moreover, they put forth a novel technique enabling nearly lossless compression of LLMs.
Nonetheless, these works fall short of achieving practical inference acceleration on edge devices, as the activation part persists in a floating-point format, rendering the integer multiplier of the edge devices ineffective.}

\subsection{Low-Bit Computation on Hardware Devices}
\py{Low-precision linear algebra kernels aim to maximize computing throughput on low-precision operands.
This is achieved by extending existing wider bit-width linear algebra kernels. The use of lower-precision operands brings about two performance enhancements: increased cache capacity and the ability to leverage lower-precision SIMD instructions for processing multiple elements simultaneously.
Pioneering examples of these low-precision linear algebra kernels, e.g., Google’s GEMMLOWP~\cite{jacob2017gemmlowp} and Facebook’s QNNPACK~\cite{dukhan2018qnnpack}, excel at enhancing the efficiency of DNN inference when employing 8-bit quantization. However, pushing for more aggressive sub-byte quantization yields no added performance benefits due to the fact that mainstream CPUs solely support SIMD operations with a precision of 8 bits or wider. In specific, low-precision kernels essentially expand sub-byte operands to 8 bits and process them accordingly.
Furthermore, the concept of Bit-serial computation emerges as a promising solution for data-parallel computation with sub-byte values. This approach involves sequentially processing each bit of two operands during multiplication, while simultaneously managing multiple operand pairs in parallel. Nonetheless, its practical implementation necessitates the \textit{\textbf{popcount}} operation, which inherently limits runtime throughput. As a result, this method only presents significant advantages in ultra-low-bit scenarios (1 or 2 bits).
%As the bit count gradually increases, the benefits of this approach rapidly diminish.
}
% Low-precision linear algebra kernels aim to maximize computing throughput on low-precision operands. This is achieved by extending existing wider bit-width linear algebra kernels. Lower-precision operands offer two performance improvements: they allow for larger cache capacity and enable lower-precision SIMD instructions to process more elements simultaneously.
% Leading low-precision linear algebra kernels like Google's GEMMLOWP are designed to optimize the throughput of low-precision operands. They work exceptionally well in improving the efficiency of DNN inference under 8-bit quantization. However, more aggressive sub-byte quantization does not provide additional performance benefits since commodity CPUs only support SIMD operations of 8-bit or wider precision. Consequently, low-precision kernels simply extend sub-byte operands to 8-bit and process them accordingly. Additionally, Bit-serial computation emerges as a promising solution for sub-byte data-parallel computation. This method involves sequentially processing each bit of two operands during multiplication, while simultaneously dealing with multiple operand pairs in parallel.
% % It offers a theoretical speedup ratio that is inversely proportional to the bit-width of the operands.
% However, its practical implementation requires \textit{\textbf{popcount}} operation, which inevitably limits the runtime throughput. Thus, such method only demonstrates sufficient advantages in ultra-low-bit (1, 2-bit) scenarios. As the number of bits gradually increases, the advantages of this method will quickly dissipate.

\section{Activation Analysis of LLMs}
In this section, we analyze the attentivity of tokens in LLMs and the influence of token pruning on activation quantization.
Besides, we deliver the latency profiling to analyze potential quantization strategy.

\begin{figure}[t]
  \centering
  \includegraphics[width=1.0\linewidth]{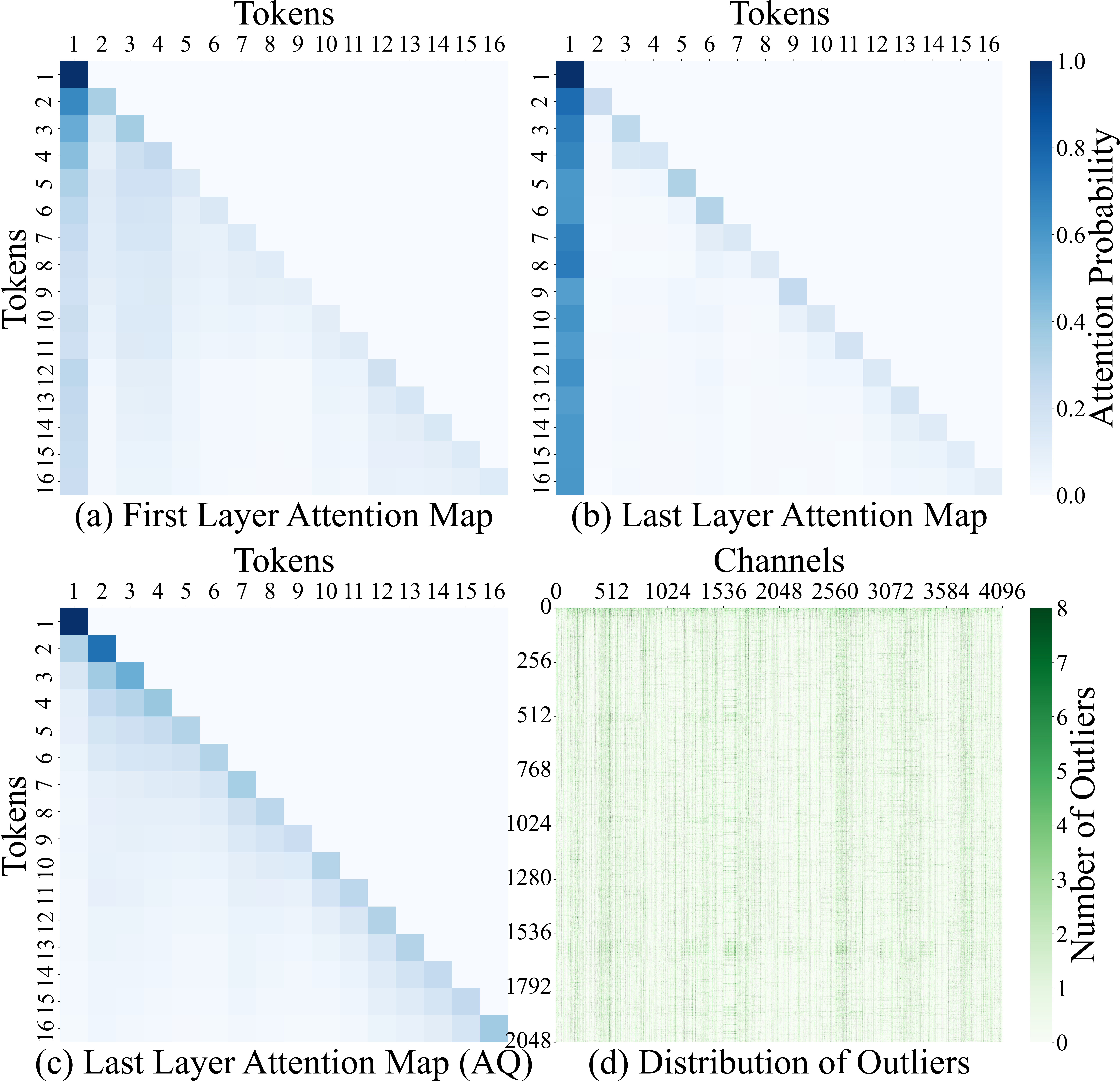}
  \caption{
  The (a), (b), and (c) shows attention maps with 16 tokens in the first and last layer of the model.
  The activation is not quantized in (a) and (b), while it is quantized in (c).
  % Several tokens in the first layer share a triangular pattern, indicating that tokens tend to the adjacent, especially the previous position.
  % Nearly all tokens in the last layer share a vertical-stripe pattern, showing that tokens all tend to the first token.
  The (d) shows the distribution of outliers in one activation with 2048 tokens.
  The visualization is based on the LLaMA-7B model with the Wikitext-2 dataset.
  }
  \label{fig:attnprob_heatmap}
\end{figure}

\subsection{Token Importance in LLMs}
\py{In natural language processing, numerous non-essential words often exist within sentences, contributing little to the overall comprehension. This implies that we can efficiently process these words using fewer resources, potentially even excluding them, in order to mitigate complexity.}
As words are embedded into tokens in language models, we explore the attention mechanism to analyze the importance of each token.
The previous works~\cite{kong2022peeling, dong2023heatvit} focus on the attention map in the transformer architectures.
The attention probabilities are then accumulated across multiple rounds of attention as token importance scores.
%, which are widely adopted in the token pruning works~\cite{wang2021spatten, berg2021keyword}.
However, the causal attention masks used in LLMs ensure that, during the self-attention mechanism, each token can only interact with previous tokens instead of the following ones.
\py{Thus, this causal mechanism makes the accumulated probabilities not appropriate to the evaluation of token importance because of its unfair for the accumulated probabilities of former tokens.}

\py{In LLMs, a distinct start token is placed at the beginning of the input sequence.
% This alters the attention map in comparison to regular language models or vision transformers.
The start token has a role in initializing the hidden layers and defining token positions within the sequence. These aspects are vital for producing text that is both coherent and contextually meaningful.}
% The work~\cite{vig2019analyzing} analyzes the attention across all of the GPT-2 model~\cite{gpt2}.
% They find that tokens only tend to the current position at the first layer and most tokens tend to the first token in the following layers.
To explore the relationship between the first start token and other tokens, we visualize the attention map at the first and last layer of the LLaMA-7B model with 16 tokens on the Wikitext-2 dataset in Figure~\ref{fig:attnprob_heatmap} (a) and (b).
\py{According to the attention map, several tokens in the first layer demonstrate a shared triangular pattern, indicating that tokens tend to the adjacent positions, especially the previous position.}
While in the last layer, nearly all tokens share a vertical-stripe pattern, indicating that tokens all related with the first token.
Then we explore the attention maps in the middle layers, showing that these maps are similar to the one in the last layer.
Thus, it guides us to build the connection between the token importance and token attentivity to the start token.

% xuan: @peiyan, talk about the quantization error
\subsection{Influence of Activation Quantization}  % Token Pruning
% xuan_todo: in the quantization, the model performance is bad, the preceding tokens can not have enough attention to the first token, which causes bad generation results.
% token pruning can reduce the distance between the preceding tokens to the start token, which can help get better performance.

% We analyze the distribution of outliers and visualize outliers of one activation in different channels in Figure~\ref{fig:attnprob_heatmap} (d).
% We find that the outliers are distributed in adjacent or even the same channels.
% It is because there are several straight lines with deep colors on the map indicating that the index of the channel containing outliers does not change.
% Also, the attention map, which is generated by the query matrix \textit{Q} and key matrix \textit{K}, can be influenced by the activation quantization as it is input dependent.
% We visualize the quantized attention map at the last layer in Figure~\ref{fig:attnprob_heatmap} (c).
% We can find that the attention map shows a triangular pattern and the quantized tokens attend to the adjacent positions rather than the first start token, which means the attention range becomes locality and the attentivity is much weaker than before.
% This kind of change indicates that the model can only generate the tokens based on some previous adjacent tokens rather than the full sequence of tokens starting from the first start token, which means the task performance becomes worse because of the activation quantization.

We analyze the distribution of outliers and visualize outlier in different channels in Figure~\ref{fig:attnprob_heatmap} (d).
\py{We notice that the outliers are distributed in adjacent or even the same channels, since several straight lines with deep colors indicates that the channel index of the outliers unchange.}
Also, the attention map, which is generated by the query \textit{Q} and key matrix \textit{K}, can be influenced by the activation quantization as it is input-dependent.
We visualize the quantized attention map at the last layer in Figure~\ref{fig:attnprob_heatmap} (c).
\py{The attention map shows a triangular pattern and the quantized tokens attend to the adjacent positions rather than the start token, demonstrating that the attention range becomes locality and the attentivity turns much weaker.
This change implies a deterioration in the globality of representative features.
From another perspective, the information reduction of the original attention map caused by quantization error will impact the final task performance adversely.}

\subsection{Latency Profiling on Hardware Devices}

\begin{figure}[tb]
\centering
\includegraphics[width=1\columnwidth]{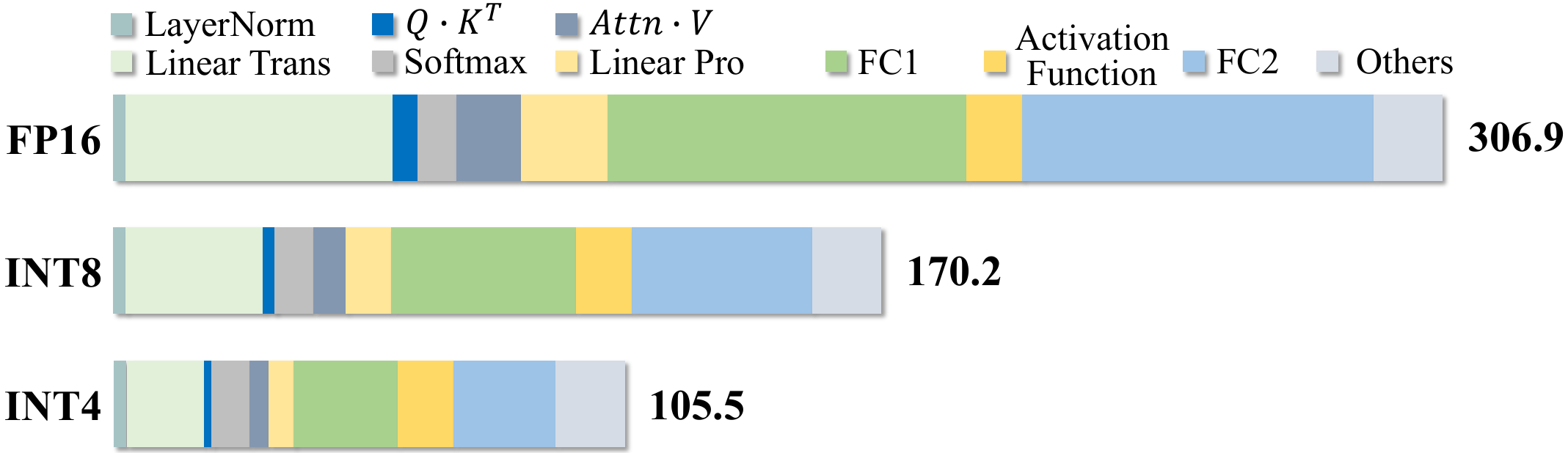}
% \vspace{-0.1in}
\caption{
Mobile Device profiling of one LLaMA block.
}
\label{fig:hardware_profile}
% \vspace{-0.2in}
\end{figure}

To gain a deeper insight into the runtime distribution of  LLMs, we conducted profiling on a widely used model, LLaMA, utilizing the on-board Snapdragon 870 CPU, as shown in Figure~\ref{fig:hardware_profile}. This profiling includes FP16, INT8, and INT4 precisions.
Since nonlinear operators (LayerNorm/Softmax/SwiGLU) contribute a relatively smaller portion of latency, i.e., $<$ 8\% for FP16, $<$ 12\% for INT8, $<$ 16\% for INT4, we have implemented them using FP16 arithmetic units to ensure task performance is maintained. We focus on the primary computation workload,  matrix multiplication, performed in various low-bit precision settings.
Our observation reveals that FC1 and FC2 account for 54\% of the runtime latency in FP16, and 49.5\% in INT8. This finding indicates the need to prioritize the quantization of these components to a lower-bit (4-bit) representation. Following that, our order of priority will be as follows: Linear Transformation $>$ Linear Projection $>$ AttnV $>$ QK.
In essence, focusing on low-bit quantization of both weights and activations for LLMs while ensuring task performance is crucial.

\section{Methodology}
We explain the activation quantization pipeline here and propose the activation-guided framework for the optimization of quantization.
Also, we explain our hardware implementation of the 4-bit multiplier.

% xuan_todo: Token pruning enhanced power-of-two quantization. inspired by (cite fqvit), we utilize the power-of-two quantization method for optimization. Meanwhile, the token prune will really prune the token (remove it), removing the token can also remove some outliers, which is helpful for power-of-two quantization because it can reduce the search space of the scales in it. talk more about the contribution of token pruning to the removal of outliers.

\subsection{Preliminary}
We here explain the quantizers we use for activation quantization in this work.
We assume the bit-width used in quantization is $b$, and then the quantizer can be defined as a function $Q(X|b)$ which can map the floating points in vector $X \in \mathbb{R}^{mxn}$ to the closest quantization in $q$:
\begin{equation} \footnotesize
    q = \left\{
             \begin{array}{lr}
             \{-2^{b-1}, ..., 2^{b-1}-1\}, & \text{Signed} \\
             \{0,1,...,2^b-1\}, & \text{Unsigned} 
             \end{array}
\right.
\end{equation}

% In our work, we mainly adopt symmetrical quantization.
There are various kinds of quantizers $Q(X|b)$, and the uniform quantizer~\cite{uniformquantizer} and the log2 quantizer~\cite{log2quant} are widely used.
In our work, we mainly use these two quantizers for activation quantization.

\noindent\textbf{Unifrom Quantization} has been supported by most hardware devices.
\begin{equation} \footnotesize
    Q(X|b) = \text{CLIP}( \lfloor \frac{X}{s} \rceil + zp , 0, 2^b-1)
\end{equation}
The $s$ and $zp$ denote the scale and zero-point separately.
% , which are determined by the following:
% \begin{equation} \footnotesize
%     s = \frac{\text{max}(X)-\text{min}(X)}{2^b-1}
% \end{equation}
% \begin{equation} \footnotesize
%     zp = \text{CLIP}(-\frac{\text{min}(X)}{s}, 0, 2^b-1)
% \end{equation}

\noindent\textbf{Log2 Quantization} imports the exponential operation into the linear quantization process.
\begin{equation} \footnotesize
    Q(X|b) = \text{Sign(}X) \cdot \text{CLIP}( \lfloor -\text{log}_2  \frac{X}{\text{max}(|X|)} \rceil , 0, 2^{b-1}-1 )
\end{equation}

\subsection{Activation Quantization Pipeline}

% 08/10/2023
% ===============================================================
% Based on the analysis in the previous section, we adopt channel-wise quantization for the optimization of the most activations.
% The channel-wise quantization can give more effective bits for the most values in those normal channels that contain even no outliers.
% For those abnormal values, we use relatively higher bits to quantize so that we can still maintain the model performance by magnifying the scales for those outliers.
% We use symmetrical quantization plan, in which the zero point is always 0, for the channel-wise quantization of LLMs, and we only need to save the scales for different channels.
% ===============================================================

\begin{figure}[t]
  \centering
  \includegraphics[width=1.0\linewidth]{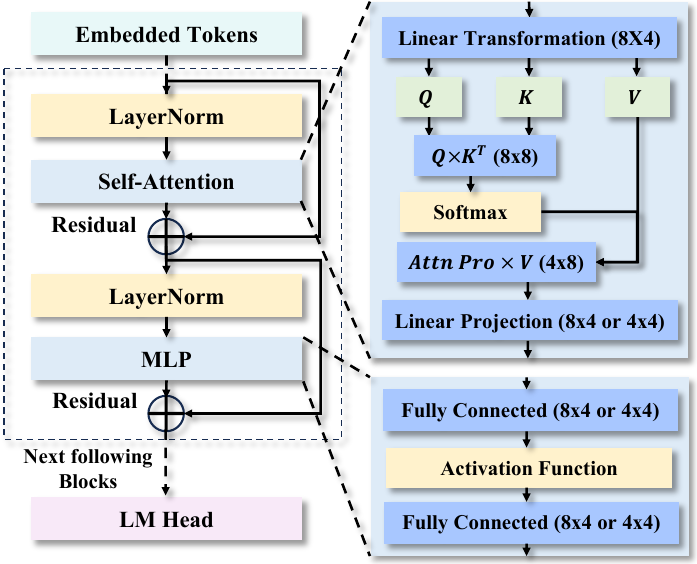}
  \caption{
  Activation Quantization Pipeline.
  % The blue modules indicate that the matrix multiplication inside can be accelerated.
  % The yellow modules are the nonlinear modules whose computation remains not changed.
  }
  \label{fig:quantization_pipeline}
\end{figure}

% We show our activation quantization pipeline in Figure~\ref{fig:quantization_pipeline}.
% As the embedding process and the output module, as well as the nonlinear operations colored yellow, do not take many proportions during model inference, we retain their computation not change.
% We focus on the acceleration of matrix multiplications which are the most proportions of the inference latency.
% In our pipeline, those computations in the self-attention module and MLP module, which are colored blue, can be accelerated by our proposed 4-bit multiplier.
% Specifically, we quantize the activations mainly with 8-bit integers, and we find that some activations after the self-attention mechanism can be quantized with 4-bit integers for further acceleration while still maintaining the task performance.

We present our activation quantization pipeline in Figure~\ref{fig:quantization_pipeline}. While the embedding process, output module, and the yellow-highlighted nonlinear operations contribute relatively small proportions during model inference, we preserve their computation without alteration. Our primary focus is optimizing the matrix multiplication operations, constituting the largest share of the inference latency.

Within our pipeline, we target the acceleration of computations occurring in the self-attention and MLP modules, as indicated by the blue shading. Specifically, we perform activation quantization predominantly using 8-bit integers. However, we observe that specific activations following the self-attention mechanism can be quantized using 4-bit integers, resulting in further acceleration while upholding task performance standards. Accordingly, we propose our innovative 4-bit multiplier to effectively support the INT4 matrix multiplication.

\subsection{Activation-Guided Optimization}

\begin{figure}[t]
  \centering
  \includegraphics[width=1.0\linewidth]{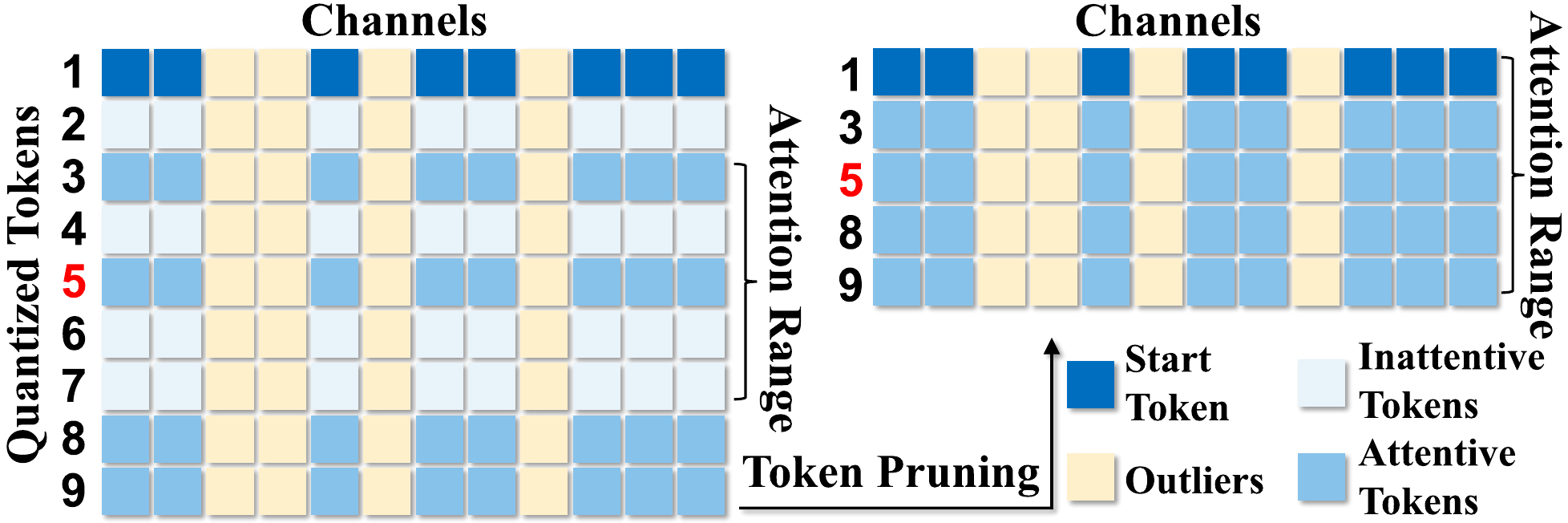}
  \caption{
  Activation Quantization With Token Pruning.
  % Activation quantization with token pruning.
  % The number of outliers is reduced and the 5th token can interact with more attentive tokens after token pruning.
  }
  \label{fig:tokenpruning}
\end{figure}

% Based on the analysis discussed in the previous section, we evaluate the token importance according to the attentivity of the token to the first start token.
% It is because the performance of the generation is highly related to the initial start token in GPT models.
% As the quantized model tends to generate tokens based on the adjacent positions, we adopt token pruning into our optimization framework to remove those redundant inattentive tokens near the important tokens and reduce the distance to the start token.
Based on the analysis of outliers and attention range in the previous section, it is intuitive for us to import the token pruning here for optimization.
We visualize the token pruning process in Figure~\ref{fig:tokenpruning}.
Token pruning can reduce the outliers, which can decrease the quantization error caused by them.
Token pruning can also reduce the distance between attentive tokens to help the model capture more features.

We first introduce the activation-aware token pruning improved activation quantization method we use in this work.
Inspired by the work~\cite{lin2022fqvit}, we propose the Two-Refine Improved by Pruning (TRIP) method here to address the difficulties in activation quantization.

For the one activation in transformer-based models, we assume it as $X \in \mathbb{R}^{m \times d}$, and the $m, d$ denote the number of tokens, and dimension separately.
We assume the token pruning function, which prunes tokens in cascade according to the token importance, as $F^P(\cdot)$.
\begin{equation} \footnotesize
    X^P = F^P( X ) \in \mathbb{R}^{ n \times d }, n < m
\end{equation}

Then, the TRIP factor $\alpha = \left\{\alpha_1, \alpha_2, ..., \alpha_d\right\}$ are applied to the different channels whose number of outliers is reduced by token pruning.
The factor $\alpha$ can provide different channels with different factors, which can regularize the quantization parameters and is utilized as:
% \zg{add $\alpha_c$ here, keep consistent with equation (8)}
\begin{equation}\footnotesize
    X{^P_Q} = Q(X^P|b, \alpha) = \text{CLIP}( \lfloor \frac{X^P}{ 2^{\alpha} s } \rceil + zp, 0, 2^b -1 )
\end{equation}
\begin{equation}\footnotesize
    s = \frac{\text{max}(X^P)-\text{min}(X^P)}{(2^b-1) \cdot 2^{\alpha}}
\end{equation}
\begin{equation}\footnotesize
    zp =\text{CLIP}( \lfloor -\frac{\text{min} (X^P) }{2^{\alpha} \cdot s} \rceil, 0, 2^b-1 )
\end{equation}
For channel $c \in [1,d]$ with the biggest refinement $K$, the factor $\alpha_c$ is as follow:
\begin{equation} \footnotesize
    \alpha_c = \mathop{\arg\min}\limits_{ \alpha_c \in \{ 0,1,...,K \} } \left\|  X{^P_c} - \lfloor \frac{X{^P_c}}{2^{\alpha_c} \cdot s} \rceil \cdot 2^{\alpha_c} \cdot s  \right\|_2
\end{equation}
The $K$ is dependent on the channels containing relatively more outliers. Outliers can be reduced by token pruning.

% The token pruning can remove some of the outliers distributed in the same or adjacent channels and reduce the distance between attentive tokens so that they can interact with each other.
For token pruning, we evaluate the token importance based on the attention map, and we only adopt their attentivity to the first start token for the importance measuring based on our analysis.
We prune tokens in cascade at the different depths of the model progressively and dynamically.
Inspired by the previous token pruning works~\cite{kim2022learned, liang2022evit}, we design our pruning strategy by splitting the model into several stages which have different redundancy~\cite{shen2023lotteryvit}.
Meanwhile, pruning tokens from some shallow layer or even from the first layer would influence the task performance of the model.
It is because, at the first few layers, the attentivity of tokens is still locality and the larger number of tokens can help the model capture more beneficial features.
Therefore, we mainly adopt the strategy that progressively pruning the tokens starting from the layer whose attention map shows that the tokens have enough ability to capture features.
Besides, we regulate the pruning ratio in each layer to balance the trade-off between loss of information and the improvement to the quantization.

\subsection{Edge-Oriented Optimization}
This section mainly proposes the hardware implementation of our \M~framework on edge devices. We first design the SIMD-based 4-bit multiplier to support the INT4 multiplication and then introduce the efficient support of 2-refined matrix multiplication.

\begin{figure}[tb]
\centering
\includegraphics[width=1.0\columnwidth]{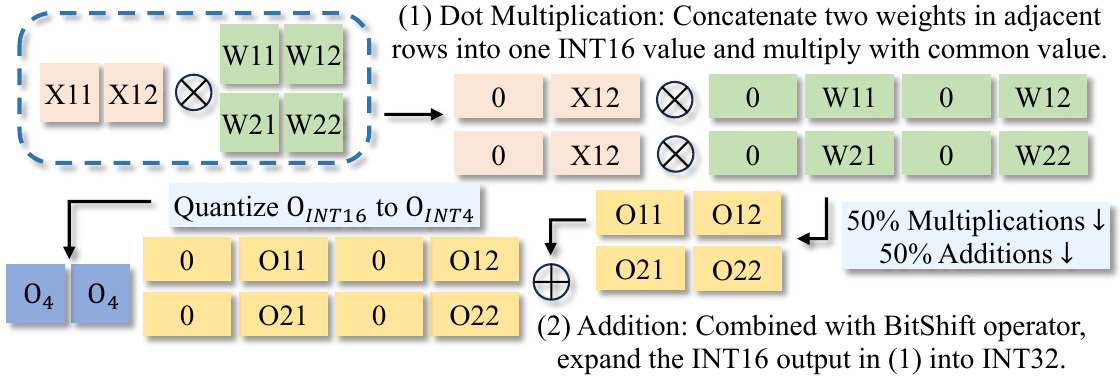}
\caption{The paradigm of INT4 multiplier.}
\label{fig:multiplier}
% \vspace{-0.5cm}
\end{figure}

\noindent\textbf{INT4 Multiplier.}
We designed a specialized 4-bit multiplier based on SIMD architecture (Figure~\ref{fig:multiplier}), aimed at supporting practical INT4 computation. Here's the workflow:
Dot-multiplication: In the SIMD kernel, we combine two adjacent weight values, $W_{i,j}$ and $W_{i+1,j}$, and multiply them with their shared activation value. The result is an INT16 data type. We allocate the first 8 bits for the multiplication with $W_{i,j}$ and the remaining 8 bits for $W_{i+1,j}$. This approach follows the SIMD memory mechanism.
Addition: By utilizing the Bitshift operator, we expand the 16-bit output from Step 1 to 32 bits. The first 8 bits are set to 0, followed by Output$_{i,j}$ in the next 8 bits, 0s in the third 8 bits, and Output$_{I+1,j}$ in the final 8 bits. We then perform a row-by-row summation. This process can handle up to $2^8$ additions without overflow, sufficient for multi-head attention (head-dimension = 32/64). Each addition has a 32-bit memory footprint.
Finally, we split the output into two INT16 values and quantize them back to INT4 at the value-level, allowing us to integrate them into the GeMM kernel.

\noindent\textbf{Efficient TRIP Matrix Multiplication.}
\begin{figure}[tb]
\centering
\includegraphics[width=1\columnwidth]{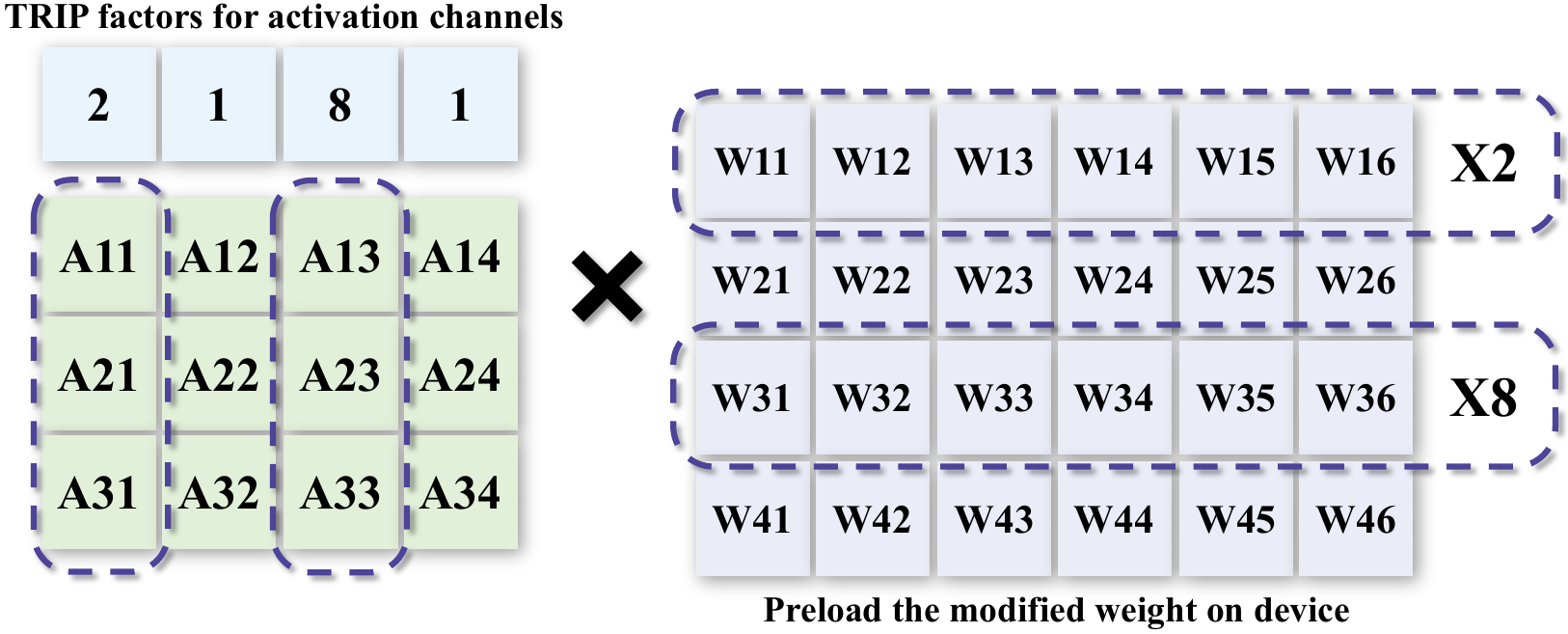}
% \vspace{-0.1in}
\caption{Hardware Implementation of Efficient TRIP Matrix Multiplication.}
\label{fig:two_mm}
% \vspace{-0.2in}
\end{figure}
Unlike channel-wise quantization, we perform layer-wise quantization on the activation matrix of outliers. 
All the channels share the same quantization parameters, i.e., scaling factors and zero-point. 
However, the predicted TRIP factors will adapt to the outlier channels' scaling factors. 
In the practical implementation, those TRIP factors will be equivalently mathematically transformed over the corresponding weights, as shown in Figure~\ref{fig:two_mm}.

Note that common inference engines only support layer-wise quantization on activation and per-channel quantization on weights, such as the GeMM and Convolution operator configuration. For example, ArmComputeLibrary~\cite{armlibrary} only supports channel-wise quantization configuration for weight matrix instead of input activation.

\section{Experiments and Results}
We introduce the experiments and the results in this section to verify the effectiveness and efficiency of our method.

\subsection{Experiment Setup}
\noindent\textbf{Setup for Activation-guided Quantization.}
We implement the activation quantization based on the weight-only quantization work GPTQ~\cite{frantar-gptq} which achieves state-of-the-art performance with 4-bit weight-only quantization for LLMs.
We mainly use 4-bit and 8-bit integers in our activation quantization.
We use the Log2 quantization for softmax activation quantization and use our TRIP quantization for other activations.
We implement the different scales of LLaMA, OPT, and BLOOM models in our experiments on the Wikitext-2 dataset~\cite{wikitextdataset} and C4~\cite{c4dataset} dataset.
% Besides, the Penn Treebank (PTB)~\cite{ptbdataset} is also used in our experiments.

\noindent\textbf{Hardware Platform.} We test the actual inference implementation on various edge devices, including the RealmeGT Android Phone with Snapdragon 870 SoC and Raspberry4 B with Quad-core CPU and 8GB RAM. Our inference engine for Arm processors is modified based on ArmComputeLibrary v22.05. The inference latency is reported via the average of 50 iterations for each test.

\subsection{Regulation of Token Pruning}

\begin{figure}[h]
  \centering
  \includegraphics[width=0.9\linewidth]{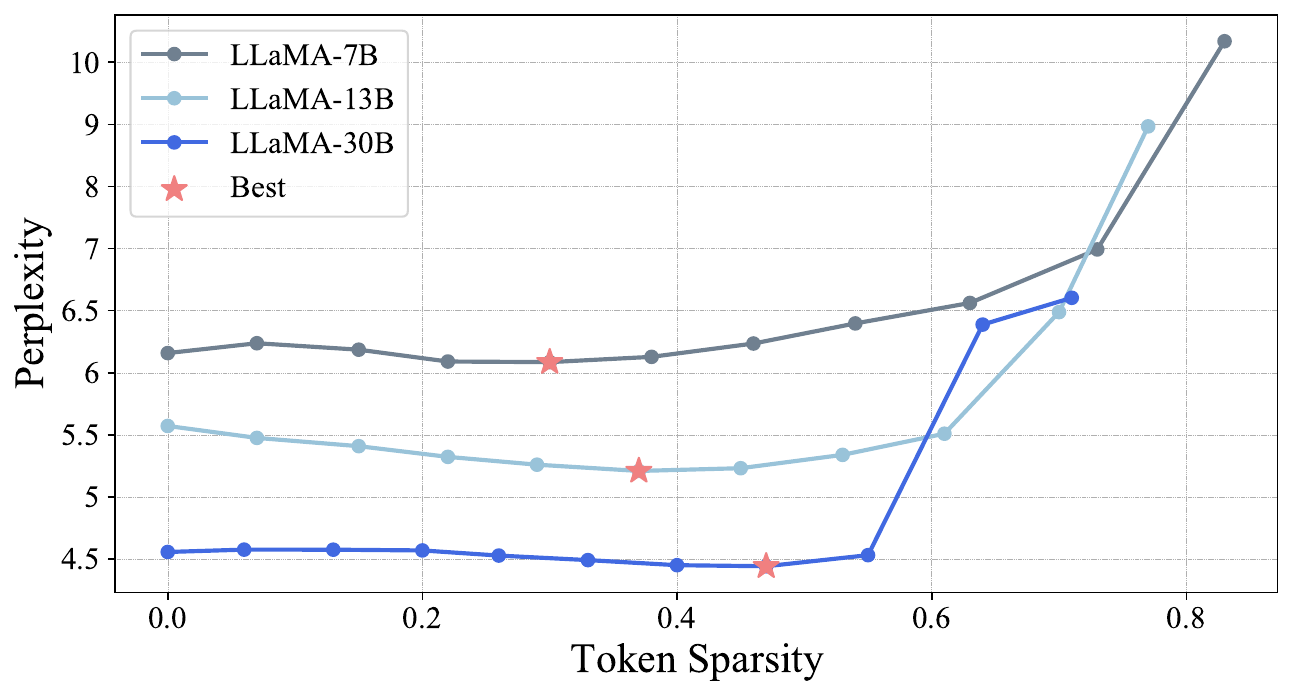}
  \caption{
  Token Sparsity vs. Perplexity.
  The visualization is based on LLaMA with 7B, 13B, and 30B scales on the Wikitext-2 dataset.
  }
  \label{fig:llama_different_size}
\end{figure}

We regulate the token pruning ratio to optimize the task performance of LLMs.
We apply the prune ratio progressively starting from the shallow layers so that the token pruning can optimize the activation quantization for more deep layers.
Assume the model has $n$ layers with $L=(l_1,l_2,...,l_n)$ and the pruning operation is added to the layers $L_p=(l_{p1},l_{p2},...,l_{pm})$.
We set prune ratio $\beta$ at the last layer $l_n$ and compute the progressive ratio $\gamma$ for the layers $l_{i} \in L_p$ as:
\begin{equation}\footnotesize
    \gamma = 1 - (1 - \beta)^{\frac{1}{m}}
\end{equation}

We accumulate the token sparsity $s$ with the number of pruned tokens during inference as:
\begin{equation}\footnotesize
    s = 1 - \sum_{i=1}^{n} r_i
\end{equation}
The $r_i$ denotes the number of remaining tokens in $l_i$.

We then adopt the weight and activation both quantized LLaMA models with 7B, 13B, and 30B to search for the optimal prune ratio.
We visualize our results in Figure~\ref{fig:llama_different_size}.
For all three different scales of LLaMA models, we use token pruning to achieve better performance than dense models at the red star points.
Also, we find that token pruning can only help the quantization achieve better results when the token sparsity is small, while token pruning makes a negative impact on the task performance when the token sparsity becomes too large.
Here we adopt the optimal token sparsity for different scales of the LLaMA model and show the exact results in Table~\ref{tab:results_llama_activation_quant}.
Also, we regulate the token pruning with the same strategy as LLaMA for OPT and BLOOM models, and the best results are shown in Table~\ref{tab:results_opt_activation_quant} and Table~\ref{tab:results_bloom_activation_quant}.

\subsection{Quantization Results and Analysis}

\begin{table}[h]
  \centering
  \resizebox{0.95\linewidth}{!}{
  \begin{tabular}{c|c c|c|c|c|c}
    \toprule
    % \multirow{2}*{Method} & Weight & Activation & Edge & \multicolumn{4}{c}{LLaMA} \\
    % \cline{5-8}
    % ~                   & Quantization & Quantization & Friendly & 7B        & 13B & 33B & 65B \\
    
    % \multirow{2}*{Method} & \multirow{2}*{WQ} & \multirow{2}*{AQ} & Edge & \multicolumn{4}{c}{LLaMA} \\
    % \cline{5-8}
    % ~                   & ~ & ~ & Friendly & 7B        & 13B & 33B & 65B \\

    \multirow{2}*{Method} & WQ & AQ  & \multicolumn{4}{c}{PPL of LLaMA} \\
    \cline{4-7}
    ~                   & \# Bits & \# Bits  & 7B        & 13B & 30B & 65B \\
    
    \midrule
    -             & FP16  & FP16 & 5.68      & 5.09 & 4.10 & 3.53\\
    RTN           & INT4  & FP16 & 6.29      & 5.53 & 4.54 & 3.92 \\
    GPTQ          & INT4   & FP16 & 5.85     & 5.2 & 4.23 & 3.65 \\
    SqLLM    & 4.05$^\ddagger$   & FP16 & 5.79     & 5.18 & 4.22 & - \\
    SpQR    & 3.94$^\ddagger$   & FP16 & 5.87     & 5.22 & 4.25 & 3.90 \\
    MoFQ8         & FP8$^\dagger$ & FP8$^\dagger$ & 6.49 & 5.41 & 5.31 & - \\
    ZQFP  & INT8   & INT8  & 5.72     &   5.09 & 4.10 & -  \\
    ZQFP  & INT4   & INT8  & 6.44     & 5.32 & 4.36 & - \\
    ZQFP$^\S$  & INT4   & INT8  & 5.88     & 5.28 & 4.34 & - \\
    \midrule
    % Ours          & INT4   & INT8  & \checkmark & 6.16      & 5.57  & 4.55  & 4.01 \\
    % Ours$^*$      & INT4   & INT8  & \checkmark & 6.09      & 5.21  & 4.44  & 3.92 \\ 
    AgileQ          & \multicolumn{2}{c|}{AgileQ-8}   & 6.16      & 5.57  & 4.55  & 4.01 \\
    AgileQ$^*$      & \multicolumn{2}{c|}{AgileQ-8}   & 6.09      & 5.21  & 4.44  & 3.92 \\ 
    
    \bottomrule
  \end{tabular}}
  \caption{
  LLaMA Quantization Results on Wikitext-2 dataset.
  AgileQ-8 denotes the 8-bit is used.
  SqLLM denotes SqueezeLLM.
  $^*$ denotes the token pruning optimized results.
  $^\dagger$ denotes the mix precision with mainly FP8 and INT8.
  $^\ddagger$ denotes the average bits.
  ZQFP denotes ZeroQuant-FP.
  $^\S$ denotes the LoRC.
  }
  \label{tab:results_llama_activation_quant}
\end{table}

% \begin{figure}[t]
%   \centering
% \includegraphics[width=0.9\linewidth]{figures/token_sparse_ppl.png}
%   \caption{
%   Activation Quantization Pipeline.
%   }
%   \label{fig:token_sparsity_ppl}
% \end{figure}

We first show the quantization results of LLaMA in Table~\ref{tab:results_llama_activation_quant}.
According to the results of weight-only quantization works, our method achieves a minor task performance drop, and we achieve better performance than most of the other activation quantization works.
The data in the last row denotes the results achieved by token pruning in our method, which verifies that the token pruning can optimize the quantization.
% Meanwhile, our work achieves better task performance with W4A8 on the Wikitext-2 dataset than MoFQ8~\cite{zhang2023integer} that uses the mix precision with mainly FP8 and INT8 in weight and activation quantization.
% % and the MoFQ8 which use mixed precision can not be implemented on the edge for inference acceleration.
% Also, our results are better than the work ZeroQuant-FP~\cite{zeroquantfp} on the scales of 7B and 13B without LoRC enhancement.
% While, the Low-Rank Compensation (LoRC) enhancement denoted as $^\S$ would increase the model size and employs low-rank matrix factorization on the quantization error matrix.
% Note that the low-rank matrix factorization can not be well-supported in the common inference engine, such as ArmComputeLibrary.
Our method achieves better task performance than MoFQ8~\cite{zhang2023integer} and ZeroQuant-FP~\cite{zeroquantfp}.
the Low-Rank Compensation (LoRC) enhancement denoted as $^\S$ would increase the model size and flops, which is also not well-supported on the common inference engine.
Then, we show our quantization result of OPT and BLOOM on the Wikitext-2 dataset and C4 dataset in Table~\ref{tab:results_opt_activation_quant} and Table~\ref{tab:results_bloom_activation_quant}.
Our method achieve better task performance than those activation quantization works and our results are close to the weight-only quantization works.
Meanwhile, token pruning also helps us achieve better task performance in activation quantization.
Especially, for OPT and BLOOM models, our method can even achieve even better task performance than the FP16 models on the C4 dataset.

\begin{table*}[t]
  \centering
  % \setlength{\leftskip}{-30pt}
  % \resizebox{1.1\linewidth}{!}{
  \resizebox{1.0\linewidth}{!}{
  \begin{tabular}{c|c|c|c|c|c|c|c|c|c|c|c|c|c}
    \toprule
    % \multirow{2}*{Method} & Weight & Activation & Edge & \multicolumn{8}{c|}{OPT on Wikitext-2} & \multicolumn{8}{c}{OPT on C4} \\
    % \cline{5-20}
    % ~                   & Quantization & Quantization & Friendly & 125M    & 350M & 1.3B & 2.7B & 6.7B & 13B & 30B & 66B & 125M    & 350M & 1.3B & 2.7B & 6.7B & 13B & 30B & 66B \\

    % \multirow{2}*{Method} & \multirow{2}*{WQ} & \multirow{2}*{AQ} & Edge & \multicolumn{8}{c|}{OPT on Wikitext-2} & \multicolumn{8}{c}{OPT on C4} \\
    % \cline{5-20}
    % ~                   & ~ & ~ & Friendly & 125M    & 350M & 1.3B & 2.7B & 6.7B & 13B & 30B & 66B & 125M    & 350M & 1.3B & 2.7B & 6.7B & 13B & 30B & 66B \\
    
    \multirow{2}*{Method} & W / A & \multicolumn{6}{c|}{PPL of OPT on Wikitext-2} & \multicolumn{6}{c}{PPL of OPT on C4} \\
    \cline{2-14}
    ~                   & \# Bits & 125M     & 1.3B & 2.7B & 6.7B & 13B & 30B  & 125M     & 1.3B & 2.7B & 6.7B & 13B & 30B  \\
    
    \midrule
    -             & 16  & 27.65  & 14.63 & 12.47 & 10.86 & 10.13 & 9.56  & 26.56  & 16.07 & 14.34 & 12.71 & 12.06 & 11.44 \\
    RTN           & \textbf{4}/16  & 37.28  & 48.17 & 16.92 & 12.10 & 11.32 & 10.98  & 33.91  & 24.51 & 18.43 & 14.36 & 13.36 & 13.46  \\
    GPTQ          & \textbf{4}/16  & 31.12  & 15.47 & 12.87 & 11.39 & 10.31 & 9.63  & 29.22 & 16.97 & 15.00 & 13.18 & 12.26 & 11.57  \\
    AWQ          & \textbf{4}/16  & 33.96  & 16.85 & 14.61 & 12.44 & 11.60 & 10.75  & -  & - & - & - & - & -   \\
    % SpQR          & 4.27$^\ddagger$   / FP16  & - & - & - & - & 10.81 & 10.22 & 9.50  & - & - & - & - & 11.88 & 11.27 & 10.73   \\
    MoFQ8         & 8$^\dagger$ / 8$^\dagger$  & -   & 16.78 & 14.24 & 12.41 & 12.52 & 10.95  & - & - & - & -  & - & - \\
    ZQV2  & \textbf{4}/\textbf{16}   & 36.71  & 19.38 & 17.92 & 11.91 & 10.67 & 10.10   & 30.92 & 17.93 & 18.32 & 13.01 & 12.07 & 11.33  \\
    % ZQFP  & INT8   & INT8   & - & - & - & 11.20 & 12.12 & 14.63  & - & -& - & - & - & 12.48 & 15.86 & 29.74   \\
    % ZQFP  & INT4   & INT8   & - & - & - & 11.61 & 12.32 & 14.80  & - & -& - & - & - & 12.92 & 16.56 & 28.09   \\
    % ZQFP$^\S$  & INT4   & INT8    & - & - & - & 11.37 & 12.06 & 15.94  & - & - & - & - & - & 12.53 & 15.85 & 31.16   \\
    \midrule
    % Ours          & INT4   & INT8  & \checkmark & 31.52 & 24.44 & 15.90 & 13.43 & 11.43 & 10.42  & 9.70 & 9.92 &  28.43      & 23.13  & 16.72  & 14.91 & 12.70      & 11.77  & 11.14  & 10.97 \\
    % Ours$^*$      & INT4   & INT8  & \checkmark &  30.37      & 24.15  & 14.90  & 13.19 & 11.21      & 10.00  & xxx  & xxx &  24.44      & 22.80  & 15.95  & 14.20 & 12.39      & 11.31  & xxx  & xxx \\
    AgileQ          & our-8   & 31.52  & 15.90 & 13.43 & 11.43 & 10.42  & 9.70  &  28.43        & 16.72  & 14.91 & 12.70      & 11.77  & 11.14   \\
    AgileQ$^*$      & our-8  &  30.37       & 14.90  & 13.19 & 11.21      & 10.00  & 9.45   &  24.44       & 15.95  & 14.20 & 12.39      & 11.31  & 11.20   \\ 
    
    \bottomrule
  \end{tabular}}
  \caption{
    Perplexity of OPT model on Wikitext-2 dataset and C4 dataset.
  our-8 denotes the 8-bit is used.
  The bold part denotes integer, otherwise float.
  $^*$ denotes token pruning optimized results.
  $^\dagger$ denotes mix precision with mainly FP8 and INT8.
  % $^\ddagger$ denotes average bits.
  % $^\S$ denotes LoRC.
  }
  \label{tab:results_opt_activation_quant}
\end{table*}

\begin{table*}[t]
  \centering
  % \resizebox{0.85\linewidth}{!}{
  \resizebox{0.9\linewidth}{!}{
  \begin{tabular}{c|c|c|c|c|c|c|c|c|c|c|c}
    \toprule
    % \multirow{2}*{Method} & Weight & Activation & Edge & \multicolumn{8}{c|}{OPT on Wikitext-2} & \multicolumn{8}{c}{OPT on C4} \\
    % \cline{5-20}
    % ~                   & Quantization & Quantization & Friendly & 125M    & 350M & 1.3B & 2.7B & 6.7B & 13B & 30B & 66B & 125M    & 350M & 1.3B & 2.7B & 6.7B & 13B & 30B & 66B \\

    % \multirow{2}*{Method} & \multirow{2}*{WQ} & \multirow{2}*{AQ} & Edge & \multicolumn{5}{c|}{BLOOM on Wikitext-2} & \multicolumn{5}{c}{BLOOM on C4} \\
    % \cline{5-14}
    % ~ & ~ & ~ & Friendly & 560M    & 1.1B & 1.7B & 3B & 7.1B & 560M & 1.1B & 1.7B & 3B & 7.1B \\

    \multirow{2}*{Method} & W/A & \multicolumn{5}{c|}{PPL of BLOOM on Wikitext-2} & \multicolumn{5}{c}{PPL of BLOOM on C4} \\
    \cline{2-12}
    ~ & \# Bits  & 560M    & 1.1B & 1.7B & 3B & 7.1B & 560M & 1.1B & 1.7B & 3B & 7.1B \\
    
    \midrule
    -             & 16/16 & 22.42 & 17.69 & 15.39 & 13.48 & 11.37 & 26.60  & 22.05 & 19.49 & 17.49 & 15.2 \\
    RTN           & \textbf{4} / 16  & 25.9 & 22.00 & 16.97 & 14.76 & 12.10 & 29.89  & 24.44 & 21.26 & 18.76 & 16.06 \\
    GPTQ          & \textbf{4} / 16 & 24.03 & 19.05 & 16.48 & 14.20 & 11.73 & 28.00 & 23.25 & 20.55 & 18.10 & 15.60 \\
    ZQV2  & \textbf{4} / \textbf{16} & 25.31 & 23.90 & 16.93 & 14.65 & 12.06 & 27.10 & 25.99 & 19.47 & 17.26 & 14.83 \\
    \midrule
    % 8-bit v:
    % Ours          & INT4   & INT8  & \checkmark &  24.01      & 18.82  & 16.23  & 14.05 & 11.73       & 26.39  & 21.80  & 19.18 &  16.96      & 14.70 \\
    % Ours$^*$      & INT4   & INT8  & \checkmark &  23.72      & 18.33  & 16.15  & 13.73 & 11.36      & 25.21  & 19.92  & 18.56 &  16.24      & 14.03  \\ 
    AgileQ          & our-8   &  24.01      & 18.82  & 16.23  & 14.05 & 11.73       & 26.39  & 21.80  & 19.18 &  16.96      & 14.70 \\
    AgileQ$^*$      & our-8 &  23.72      & 18.33  & 16.15  & 13.73 & 11.36      & 25.21  & 19.92  & 18.56 &  16.24      & 14.03  \\ 
    \bottomrule
  \end{tabular}}
  \caption{
    Perplexity of BLOOM model on the Wikitext-2 dataset and C4 dataset.
  our-8 denotes the 8-bit is used.
  The bold part denotes integer, otherwise float.
  $^*$ denotes the token pruning optimized results.
  ZQV2 denotes the ZeroQuant-V2
  }
  \label{tab:results_bloom_activation_quant}
\end{table*}

\subsection{Ablation Study}

\begin{figure}[b]
  \centering
  \includegraphics[width=0.9\linewidth]{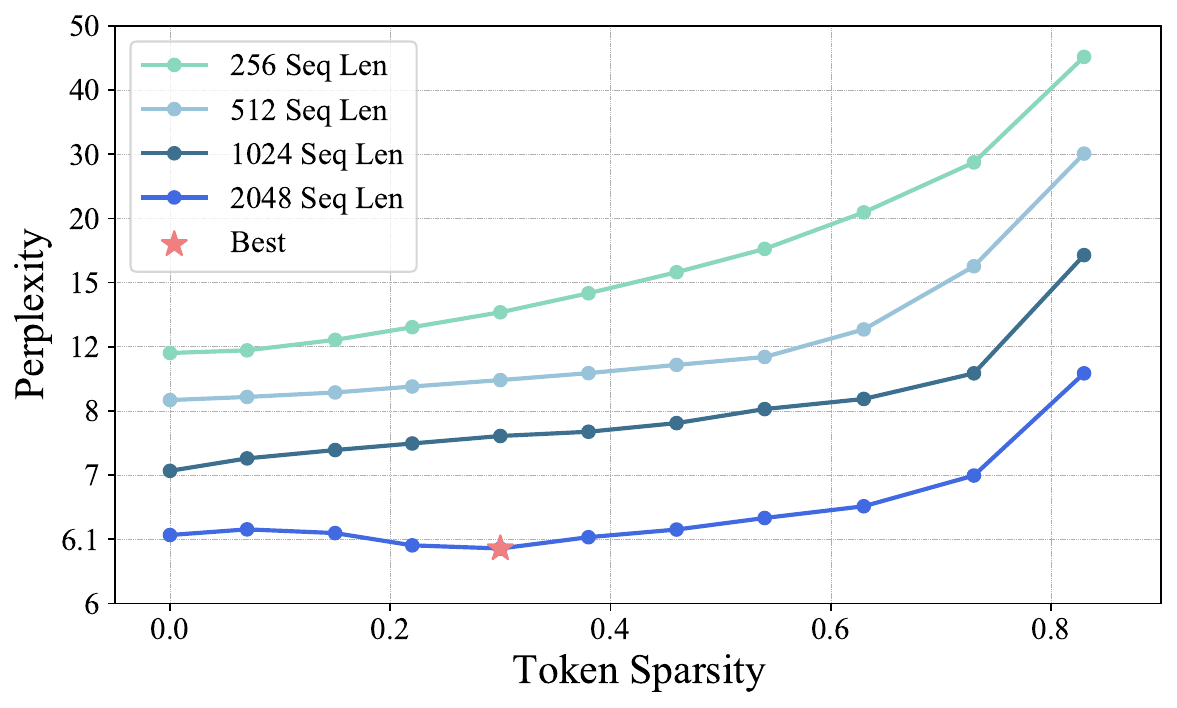}
  \caption{
  Token Sparsity vs. Input Sequence Length.
  The visualization is based on LLaMA-7B with Wikitext-2 dataset.
  }
  \label{fig:tokenpruning_inputsequence}
\end{figure}

The length of the input sequence makes a big influence on the evaluation process, we try the token pruning with different sequence lengths to further explore the variation of performance.
The normal default length of the input sequence used in the evaluation of LLMs is 2048, which is widely used in LLMs-related works.
% ~\cite{frantar-gptq, lin2023awq, dettmers2023spqr}.
Thus, we regulate the input sequence length of LLaMA-7B with Wikitext-2 dataset to explore the relationship between it and token sparsity.
The results are in Figure~\ref{fig:tokenpruning_inputsequence}.
We can find that task performance becomes worse as the sequence length becomes shorter and the token sparsity becomes larger, and the token pruning only works when the sequence length is long enough (i.e., 2048).
% The model can only be optimized by the token pruning with the longest sequence 2048.

\subsection{End-to-end Performance and Analysis}

\begin{table}[]
  \centering
  \resizebox{0.95\linewidth}{!}{
\begin{tabular}{c|c|c|c|c}
\toprule
 \multirow{2}{*}{\# Bits} & \multirow{2}{*}{\begin{tabular}[c]{@{}c@{}}Size \\ (GB)\end{tabular}} & \multirow{2}{*}{PPL} & \multicolumn{1}{c|}{Android} & \multicolumn{1}{c}{Raspberry} \\ %\cline{5-8} 
                          &                         &                                                                    &    \multicolumn{1}{c|}{CPU (s)}   &  \multicolumn{1}{c}{Pi (s)}  \\ 
                          \midrule %\hline
                          \multicolumn{5}{c}{OPT-125M} \\ \midrule
 FP16                    & 0.24  & 27.65                                                                 & 1.03              \ 1$\times$            & 143.2             \ 1$\times$       \\ 
                           ours$^*$-8                                    & 0.04 & 30.37                                                                 & 0.58              \ 1.7$\times$          & 80.34            \ 1.7$\times$        \\
                           ours$^*$-4                 		      & 0.04      & 36.95                                                            & 0.44              \ 2.3$\times$         & 61.16            \  2.3$\times$         \\ 
                          \midrule %\hline
 %                          \multicolumn{5}{c}{OPT-350M} \\ \midrule
 % FP16                    & 0.63    & 22.00                                                              & 2.79         \     1$\times$         & 387.95        \    1$\times$          \\
 %                           ours$^*$-8                		      & 0.12        & 24.15                                                          & 1.57             \ 1.8$\times$       & 218.00          \     1.7$\times$        \\
 %                           ours$^*$-4       				      & 0.12             & 26.72                                                     & 1.16            \ 2.4$\times$       & 162.14            \ 2.3$\times$      \\ 
 %                          \midrule %\hline
                          \multicolumn{5}{c}{OPT-1.3B} \\ \midrule
 FP16                      & 2.50        & 14.63                                                         & 5.42         \     1$\times$      & 754.25          \  1$\times$         \\
                           ours$^*$-8               		      & 0.49               & 14.90                                                   & 2.98               \ 1.8$\times$         & 410.58           \ 1.8$\times$     \\
                           ours$^*$-4     				      & 0.49                & 18.20                                                  & 2.18              \ 2.5$\times$        & 296.50        \    2.5$\times$    \\ 
                          \midrule %\hline
                          \multicolumn{5}{c}{OPT-2.7B} \\ \midrule
 FP16                     & 5.00          & 12.47                                                   & 8.28            \  1$\times$     & 1150.9          \ 1$\times$        \\
                           ours$^*$-8                                    & 0.94       & 13.19                                                         & 4.60           \    1.8$\times$         & 625.41            \ 1.8$\times$    \\
                           ours$^*$-4                                        & 0.94   & 16.32                                                             & 3.31          \    2.5$\times$       & 455.64            \ 2.5$\times$    \\ 
                          \midrule %\hline
                          \multicolumn{5}{c}{LLaMA-7B} \\ \midrule
 FP16                   & 13.5       & 5.68                                                       & 10.6              \ 1$\times$       & 1473.4           \ 1$\times$          \\
                           ours$^*$-8                                   & 2.53     & 6.09                                                            & 5.89              \ 1.8$\times$       & 810.25            \ 1.8$\times$       \\
                           ours$^*$-4                                       & 2.53   & 8.81                                                             & 4.44           \   2.4$\times$        & 610.98           \ 2.3$\times$         \\ \bottomrule
\end{tabular}
}
  \caption{
  Hardware results under different data precision for various LLMs.
  % ours-8 or ours-4 denotes the 8-bit or 4-bit is mainly used in \M separately.
  % $^*$ denotes the token pruning optimized results.
  % * token\_length=128.
  Results are obtained by \M~with 4 or 8 bits and token pruning.
  }
\label{tab:results_edge_devices}
\end{table}

Based on Table~\ref{tab:results_edge_devices}, 
% the following conclusions could be drawn:
we can find that \M~can bring an overall acceleration ratio of 2.3x to 2.6x depending on the model.
% as the high computational workload on edge devices can benefit from the higher utilization efficiency of limited memory resources under quantization algorithms.
% Specifically, 
Especially, Agile-Quant-8 quantizes the activation into INT8 precision and can achieve approximately 1.8x to 1.9x acceleration compared to FP16 in GeMM.
% This is very beneficial for models like transformers, where most mathematical operations are matrix operations ($>$80\% computation). 
% Hence, transformers can obtain more speedup gain than CNN, whose memory movement, such as weight reshaping, image2column, and column2image under different data layouts in each convolution operation, can dilute the efficient quantization speedup on GeMM.
Also, combined with the 4-bit compression and concatenation technique, Agile-Quant-4 can further improve this advantage, achieving approximately 1.75x acceleration compared to INT8 multiplication.
% This is because, while the theoretical, computational workload is halved, overhead is introduced due to internal shifts of concatenated weights and the recovery of stored results in INT8 format.

\section{Conclusions and Limitations}
In this paper, we propose \M, an activation-guided quantization framework for popular LLMs, and design an end-to-end accelerator on multiple edge devices.
We introduce the quantization strategy on model weights and activations, and we import token pruning to optimize quantization.
We introduce SIMD-based 4-bit multiplier and efficient TRIP matrix multiplication to achieve the 2.55x speedup on hardware devices.
% implement the accelerator for LLMs on Android CPU and Raspberry Pi, reaching up to 2.55x speedup.
Our next step is to explore lower-bit LLMs and design multiple lower-bit multipliers.
% (2,3-bits) on edge processors to proceed with AI democratization.

\section{Acknowledgements}

% This research is funded in whole or in part by the Army Research Office/Army Research Laboratory via grant W911-NF-20-1-0167 to Northeastern University. 
% Any errors and opinions are not those of the Army Research Office or Department of Defense and are attributable solely to the author(s). 
% This research is also partially supported by the National Science Foundation CCF-1937500 and CNS-1909172.

This research is mainly funded by the Army Research Office/Army Research Laboratory via grant W911-NF-20-1-0167 to Northeastern University and is partially supported by the National Science Foundation CCF-1937500 and CNS-1909172.

\bibliography{aaai24}

% \newpage
% \section{Appendix}

\begin{table*}[t]
\centering
  \resizebox{0.75\linewidth}{!}{

\begin{tabular}{cc|ccc|ccc|ccc|ccc}
\toprule
WQ               & AQ              & \multicolumn{3}{c|}{PPL of LLaMA-7B}                                         & \multicolumn{3}{c|}{PPL of LLaMA-13B}                                        & \multicolumn{3}{c|}{PPL of LLaMA-30B}                                        & \multicolumn{3}{c}{PPL of LLaMA-65B} \\ \midrule
\# Bits          & \# Bits         & \multicolumn{1}{c}{WIKI} & \multicolumn{1}{c}{C4} & \multicolumn{1}{c|}{PTB} & \multicolumn{1}{c}{WIKI} & \multicolumn{1}{c}{C4} & \multicolumn{1}{c|}{PTB} & \multicolumn{1}{c}{WIKI} & \multicolumn{1}{c}{C4} & \multicolumn{1}{c|}{PTB} & \multicolumn{1}{c}{WIKI}  & C4 & PTB \\ \midrule
FP16             & FP16            & 5.68                     & 7.08                   & 27.34                    & 5.09                     & 6.61                   & 19.23                    & 4.10                     & 5.98                   & 16.29                    & 3.53                     &  5.62  &  17.61   \\ 
INT4             & FP16            & 5.85                     & 7.23                   & 27.80                    & 5.20                     & 6.71                   & 19.87                    & 4.23                     & 6.07                   & 16.47                    & 3.65                      &  5.69  &  24.44   \\ 
\multicolumn{2}{c|}{Agile-Quant-8} & 6.16                     & 7.66                   & 29.76                    & 5.57                   & 7.39                   & 21.59                    & 4.55                     & 6.71                   & 17.23                    & 4.01                      &  6.37  &   17.35  \\ 
\multicolumn{2}{c|}{Agile-Quant-8$^*$} & 6.09                     & 7.51                   & 25.29                    & 5.21                   & 6.83                   & 12.11                    & 4.44                     & 6.61                   & 12.36                    & 3.92                      &  5.95  &  12.87   \\ 
\bottomrule
\end{tabular}
  
}

\caption{
  Full results of the LLaMA models on Wikitext-2, C4, and PTB datasets. 
  \M-8 denotes the 8-bit is used.
  $^*$ denotes the token pruning optimized results.
  }
\label{tab:results_llama_c4_ptb}

\end{table*}

\begin{table*}[h]
\centering
  \resizebox{0.85\linewidth}{!}{

\begin{tabular}{cc|cccccccc|ccccc}
\toprule
WQ               & AQ              & \multicolumn{8}{c|}{PPL of OPT on PTB}                                                                                                                                                                                            & \multicolumn{5}{c}{PPL of BLOOM on PTB}                                                                                                          \\ \midrule
\# Bits          & \# Bits         & \multicolumn{1}{c}{125M} & \multicolumn{1}{c}{350M} & \multicolumn{1}{c}{1.3B} & \multicolumn{1}{c}{2.7B} & \multicolumn{1}{c}{6.7B} & \multicolumn{1}{c}{13B} & \multicolumn{1}{c}{30B} & \multicolumn{1}{c|}{66B} & \multicolumn{1}{c}{560M} & \multicolumn{1}{c}{1.1B} & \multicolumn{1}{c}{1.7B} & \multicolumn{1}{c}{3B} & \multicolumn{1}{c}{7.1B} \\ \midrule
FP16             & FP16            & 38.99                    & 31.08                    & 20.29                    & 17.97                    & 15.77                    & 14.52                   & 14.04                   & 13.36                    & 43.69                    & 57.96                    & 30.00                       & 25.34                  & 20.83                    \\
INT4             & FP16            & 45.17                    & 34.52                    & 21.85                    & 19.14                    & 16.56                    & 14.94                   & 14.26                   & 13.81                    & 46.97                    & 62.47                    & 31.84                    & 26.49                  & 21.67                    \\
\multicolumn{2}{c|}{Agile-Quant-8} & 37.57                    & 29.33                    & 18.78                    & 16.46                    & 13.81                    & 12.78                   & 12.12                   & 12.07                    & 45.49                    & 52.15                    & 30.48                    & 24.48                  & 20.33                    \\
\multicolumn{2}{c|}{Agile-Quant-8$^*$} & 34.26                    & 27.63                    & 16.62                    & 15.98                    & 13.34                    & 12.32                   & 12.65                   & 11.62                    & 43.13                    & 57.15                    & 29.16                    & 24.11                  & 19.01                    \\ 
\bottomrule
\end{tabular}

}

\caption{
  The results of the OPT and BLOOM models on the PTB dataset. 
  \M-8 denotes the 8-bit is used.
  $^*$ denotes the token pruning optimized results.
  }
\label{tab:results_opt_bloom_ptb}

\end{table*}

\newpage

\section{Appendix}

\subsection{Additional Results}

We deliver the additional results for LLaMA models on the C4 and PTB datasets in Table~\ref{tab:results_llama_c4_ptb}, and the OPT and BLOOM models on the PTB dataset in Table~\ref{tab:results_opt_bloom_ptb}.

% \begin{algorithm}
% \caption{4-bit Multiplier\ Implementation}\label{sfm}
% \begin{lstlisting}
% 4-bit_GeMM_4x4(i, s1, s2):
% s1 = [ 4x4 matrix of src1 ]
% s2 = [ 1x4 vector of src2 ]
% c = [ 4x4 matrix of zeros ]
% mask = [ 1x4 vector of 0x00FF00FF ]
% p = [ 1x4 vector with zeros ]
% /* Inner loop for 16 elements
%    4 units per loop */
% for j in range(4):
%     /* Lane-wise Multiplication */
%     p = [multiply the j-th row of s1 with the i-th element of s2]
%     /* Product Rearrangement */
%     t = [left shift elements of p by 8 bits]
%     p = [bitwise OR between p, t]
%     p = [bitwise AND between p, mask]
%     /* Accumulation */
%     c.row[j] = [add elements of p to the corresponding j-th row of c]
% return c
        
% \end{lstlisting}
% \end{algorithm}
\subsection{The Implementation Details of 4-bit Multipliers}

\begin{algorithm}
\caption{4-bit Multiplier\ Implementation}\label{sfm}
\begin{lstlisting}
4-bit_GeMM_4x4(i, s1, s2):
s1 = [ 4x4 matrix of src1 ]
s2 = [ 1x4 vector of src2 ]
c = [ 4x4 matrix of zeros ]
mask = [ 1x4 vector of 0x00FF00FF ]
p = [ 1x4 vector with zeros ]
/* Inner loop for 16 elements
   4 units per loop */
for j in range(4):
    /* Lane-wise Multiplication */
    p = [multiply the j-th row of s1 with the i-th element of s2]
    /* Product Rearrangement */
    t = [left shift elements of p by 8 bits]
    p = [bitwise OR between p, t]
    p = [bitwise AND between p, mask]
    /* Accumulation */
    c.row[j] = [add elements of p to the corresponding j-th row of c]
return c
        
\end{lstlisting}
\end{algorithm}

 The implementation below has been programmed based on ArmISAs and tested on the maincore 3.2GHz Cortex A77 of Snapdragon 870 onboard CPU. The kernels have been implemented within ArmComputeLibrary v22.05 Inference framework.

 The breakdown of MLA (multiplication \& Addition) with MUL and ADD does not introduce any obvious inference latency differences according to our benchmarks on both Snapdragon 870 onboard-CPU and RaspberryPi 4B, opening up the possibility of inserting result-adjustment auxiliary operations between the MUL and ADD to achieve SMMW (single-multiplication-multiple-weight). The 4-bit Multiplier has been implemented based on ARM ISAs following: MUL (Multiplication), LSL (LeftShift), ORR (BitwiseOR), AND (BitwiseAND) the same process as Figure~\ref{fig:multiplier} in our paper.

\end{document}